\documentclass[11pt]{article}

\usepackage[preprint]{acl}

\usepackage{times}
\usepackage{latexsym}

\usepackage{booktabs}        
\usepackage{multirow}        
\usepackage{colortbl}        
\usepackage{xcolor}          
\usepackage{siunitx}         
\usepackage{array}           
\usepackage{subcaption} 

\usepackage{amsfonts}
\usepackage{amsmath}
\usepackage{booktabs}
\usepackage{siunitx}
\usepackage{array}

\usepackage[T1]{fontenc}

\usepackage[utf8]{inputenc}

\usepackage{microtype}

\usepackage{inconsolata}

\usepackage{graphicx}

\title{Semantics Meet Signals: Dual Codebook Representationl Learning for  Generative  Recommendation }

\author{
    \textbf{Zheng Hui}$^{1,}$,
    \textbf{Xiaokai Wei}$^{2}$,
    \textbf{Reza Shirkavand}$^{3}$,
    \textbf{Chen Wang}$^{2}$,\\
    \textbf{Weizhi Zhang}$^{4}$,
    \textbf{Alejandro Peláez}$^{2}$,
    \textbf{Michelle Gong}$^{2}$ \\
    $^{1}$University of Cambridge 
    $^{2}$Roblox Corporation \\
    $^{3}$University of Maryland 
    $^{4}$University of Illinois Chicago \\
    \texttt{zh2483@columbia.edu} 
    \texttt{\{xwei, cwang, apelaez, mgong\}@roblox.com} \\
    \texttt{rezashkv@cs.umd.edu, wzhan42@uic.edu}
}

\begin{document}
\maketitle
\begin{abstract}
Generative recommendation has recently emerged as a powerful paradigm that unifies retrieval and generation, representing items as discrete semantic tokens and enabling flexible sequence modeling with autoregressive models. Despite its success, existing approaches rely on a single, uniform codebook to encode all items, overlooking the inherent imbalance between popular items rich in collaborative signals and long-tail items that depend on semantic understanding. We argue that this uniform treatment limits representational efficiency and hinders generalization. To address this, we introduce FlexCode, a popularity-aware framework that adaptively allocates a fixed token budget between a collaborative filtering (CF) codebook and a semantic codebook. A lightweight MoE dynamically balances CF-specific precision and semantic generalization, while an alignment and smoothness objective maintains coherence across the popularity spectrum. We perform experiments on both public and industrial-scale datasets, showing that FlexCode consistently outperform strong baselines. FlexCode provides a new mechanism for token representation in generative recommenders, achieving stronger accuracy and tail robustness, and offering a new perspective on balancing memorization and generalization in token-based recommendation models.
\end{abstract}

\section{Introduction}

Recommender systems are a central component of modern information platforms, enabling personalized ranking and retrieval in e-commerce, streaming media, and social feeds \cite{10.1145/2843948, wang2025solving, hui2025matcha}. Traditional recommendation methods \cite{5197422, he2017neural, sun2019bert4rec, geng2022recommendation} learn latent representations from observed user--item interactions. These models are effective for frequently interacted (head) items because they are able to memorize fine-grained co-occurrence structure. However, they exhibit systematic degradation on long-tail and cold-start items, where interaction data is sparse \cite{zhang2021model, liu2023diffusion}. This reflects an inherent memorization bias: collaborative filtering captures relational specificity but lacks semantic generalization.
Semantic-ID based Generative recommendation \cite{rajput2023recommender,wang2024learnable} has recently emerged as a compelling alternative paradigm. Rather than treating recommendation as score prediction over fixed item IDs, generative recommenders map items into discrete semantic tokens using vector quantization or learned codebooks, and then train an autoregressive model to generate future items in sequence. This formulation offers several advantages. It unifies recommendation with sequence generation, making it compatible with large language models and multilingual or multimodal supervision. It also supports better generalization to unseen items, since tokenization can draw on textual, visual, or metadata semantics rather than relying solely on collaborative history \cite{lv2024semantic}.

Despite these benefits, current generative recommenders suffer from two fundamental limitations that prevent them from fully replacing traditional CF systems in practice. The first limitation is representation entanglement \cite{shirkavand2025catalog, fang2025hid}. Most existing approaches adopt a single, shared codebook for all items, and attempt to embed both semantic content information and collaborative interaction information into the same quantized space. This forces the model to compress heterogeneous factors of variation into a single representation. As a consequence, the learned item tokens are often neither semantically pure nor collaboration-accurate. The same codeword is expected to simultaneously reflect textual or visual meaning and encode high-order co-consumption structure. This coupling leads to interference. For head items, the semantic signal can dilute highly localized collaborative patterns. For tail items, the collaborative signal can dominate even when it is statistically unreliable, which harms cold-start behavior. In short, entanglement in a single codebook induces representational collapse: neither factor is modeled optimally \cite{baykal2024edvae}.
The second limitation is static capacity allocation \cite{zhang2024towards}. Existing semantic identifier and unified ID-plus-semantic methods allocate a fixed representational budget to each item. Every item is assigned the same number and type of codebook tokens, regardless of its popularity and data regime. This is in tension with the long-tailed nature of recommender data \cite{klimashevskaia2024survey}. Head items have abundant interaction data and therefore benefit from high-capacity collaborative representations. Tail items, on the other hand, lack interaction support and rely more strongly on semantic evidence such as text descriptions, visual attributes, or structured metadata. Treating these two classes of items identically wastes capacity in one regime and starves it in the other. In particular, it leads to overfitting on the head while under-representing the tail.

We view these two limitations as a more general problem in generative recommendation: adaptive capacity allocation. Given a fixed token budget per item, how should a system decide how much of that capacity to devote to collaborative specificity versus semantic generalization, and how can it avoid destructive interference between the two? 
To address these challenges, we propose \textbf{FlexCode}, a popularity-aware generative recommendation framework that explicitly disentangles collaborative and semantic channels and dynamically allocates representational capacity between them.

\textbf{FlexCode} introduces two separate codebooks: a collaborative codebook $\mathbf{C}_{\mathrm{CF}}$ that captures high-order interaction structure, and a semantic codebook $\mathbf{C}_{\mathrm{SEM}}$ that captures modality-grounded meaning. Our framework uses a lightweight mixture-of-experts (MoE) gating mechanism that conditions on item statistics such as popularity or interaction sparsity, and selects how much capacity to allocate to each codebook under a fixed overall token budget. Intuitively, head items are routed toward collaborative experts and receive more tokens from $\mathbf{C}_{\mathrm{CF}}$, while tail items are routed toward semantic experts and receive more tokens from $\mathbf{C}_{\mathrm{SEM}}$.
In addition to disentangling representation sources and adapting capacity, \textbf{FlexCode} introduces two regularization objectives to ensure stability and coherence. Together, these components allow \textbf{FlexCode} to interpolate smoothly between collaborative-style and semantic-style item encodings, rather than forcing a hard dichotomy. Conceptually, this approach operationalizes the intuition that memorization and generalization are not mutually exclusive, but should be emphasized differently for different regions of the item distribution.

We evaluate \textbf{FlexCode} on both publicly available benchmarks and large-scale industrial datasets. Across all settings, our method outperforms Item-ID based, Semantic-ID and unified representation baselines on standard ranking metrics.
Our main contributions are as follows:
\begin{itemize}
    \item We identify two structural limitations of existing generative recommenders: representation entanglement in a single shared codebook and static capacity allocation that ignores item popularity and data sparsity. We formalize these limitations as instances of an adaptive capacity allocation problem.
    \item We propose \textbf{FlexCode}, a framework that factorizes item representations into a collaborative codebook and a semantic codebook, and uses a popularity-aware mixture-of-experts gate to allocate a fixed token budget between them on a per-item basis.
    \item We demonstrate consistent improvements over recent Semantic-ID and unified-ID baselines on both public and industrial datasets, and show that \textbf{FlexCode} yields better head--tail balance and superior robustness under tight token budgets.
\end{itemize}

\begin{figure*}[h]
\centering
\includegraphics[width=0.9\linewidth]{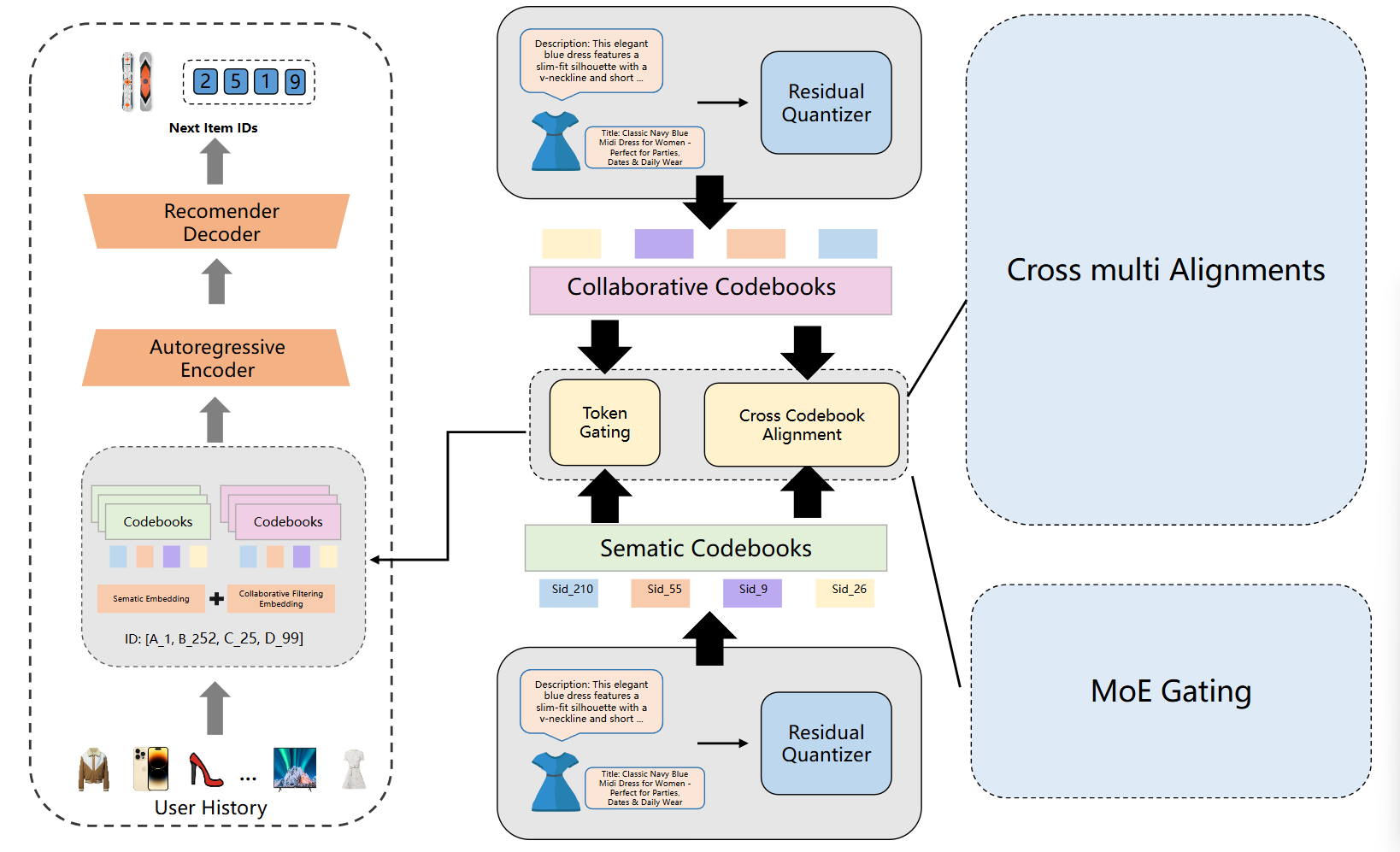}
\caption{
Overview of the \textbf{FlexCode} framework for generative recommendation.
Each item is encoded by a dual codebook with collaborative and semantic codebooks, aligned via a cross-codebook contrastive objective.
A popularity-aware Mixture-of-Experts (MoE) router adaptively allocates the budget between codebooks, and an autoregressive Transformer is trained on the resulting sequences to generate items.
}
\label{fig:main}
\end{figure*}

\section{Background and Preliminaries}

\subsection{Generative Recommendation Framework}

Let $\mathcal{U}$ denote the set of users and $\mathcal{I}$ the set of items.  
For each user $u \in \mathcal{U}$, we observe an interaction sequence 
\[
\mathbf{s}_u = [i_1, i_2, \ldots, i_T],
\]
where $i_t \in \mathcal{I}$ represents the $t$-th item interacted with by user $u$.  
The goal of recommendation is to predict the next item $i_{T+1}$ given the prefix sequence $\mathbf{s}_u$.

In \emph{generative recommendation}, this problem is reformulated as a conditional sequence generation task.  
Each item $i_t$ is first mapped into one or more discrete tokens through a vector-quantized codebook,  
\[
\mathbf{z}_{i_t} = [z_{i_t}^1, z_{i_t}^2, \ldots, z_{i_t}^{L}],
\]
where each $z_{i_t}^l \in \{1, \ldots, K\}$ indexes one entry in a learnable codebook $\mathbf{C} \in \mathbb{R}^{K \times d}$,  
$L$ is the number of tokens per item, and $d$ is the embedding dimension.  
An autoregressive model $p_\theta$ (e.g., a Transformer decoder) is then trained to maximize the likelihood of the item sequence in tokenized form:
\[
\mathcal{L}_{\text{gen}} = - \sum_{u \in \mathcal{U}} \sum_{t=1}^{T} \log p_\theta(\mathbf{z}_{i_t} \mid \mathbf{z}_{i_{<t}}).
\]
This formulation allows recommendation to be cast as a language modeling problem, enabling the use of large pre-trained generative models.

\subsection{Collaborative and Semantic Representations}

Traditional collaborative filtering (CF) models learn an embedding matrix 
$\mathbf{E}_{\text{CF}} \in \mathbb{R}^{|\mathcal{I}| \times d}$ directly from user--item interactions.  
Given user embedding $\mathbf{e}_u$ and item embedding $\mathbf{e}_i$, the relevance score is typically computed as
\[
r(u, i) = \mathbf{e}_u^\top \mathbf{e}_i,
\]
and optimized with implicit feedback losses such as Bayesian Personalized Ranking (BPR).

By contrast, semantic encoders (e.g., textual or visual models) map content features $\mathbf{x}_i$ into a representation
\[
\mathbf{e}_{\text{sem}}(i) = f_{\text{sem}}(\mathbf{x}_i),
\]
capturing modality-derived meaning independent of user interactions.  
While semantic representations generalize better to cold-start items, they lack collaborative precision.

Generative recommenders based on semantic identifiers (Semantic-IDs) \citep{rajput2023recommender} quantize these embeddings into discrete tokens via vector quantization:
\[
\hat{z}_i = \arg\min_{k} \| \mathbf{e}_{\text{sem}}(i) - \mathbf{C}_k \|_2^2,
\]
producing token indices that are used for sequence generation.  
However, existing methods employ a single codebook $\mathbf{C}$ for all items, mixing collaborative and semantic signals implicitly.  
This \emph{uniform codebook assumption} limits representational flexibility and ignores item popularity heterogeneity.

\subsection{Problem Formulation: Capacity Allocation}

Let $\mathbf{C}_{\text{CF}}$ and $\mathbf{C}_{\text{SEM}}$ denote the collaborative and semantic codebooks, respectively.  
Given a fixed total token budget $L$, we define a function
\[
L_{\text{CF}}(i) + L_{\text{SEM}}(i) = L,
\]
where $L_{\text{CF}}(i)$ and $L_{\text{SEM}}(i)$ determine how many tokens of item $i$ are drawn from each codebook.  
The objective of capacity allocation is to learn an adaptive mapping
\[
g(i) = \sigma(w \cdot \log(\text{pop}(i)) + b),
\]
where $\text{pop}(i)$ is the empirical popularity (interaction frequency) of item $i$,  
and $g(i) \in [0,1]$ determines the proportion of CF vs. semantic tokens.  
Accordingly,
\[
L_{\text{CF}}(i) = \lfloor g(i) \cdot L \rfloor, \quad 
L_{\text{SEM}}(i) = L - L_{\text{CF}}(i).
\]
For head items (high $\text{pop}(i)$), $g(i)$ approaches 1, allocating more CF tokens;  
for tail items, $g(i)$ decreases, emphasizing semantic tokens.  
This defines an \emph{adaptive dual-codebook encoding} that allocates capacity according to item popularity.

The resulting generation objective combines both token sources:
\[
\mathcal{L}_{\text{gen}}^{\text{dual}} = 
- \sum_{u \in \mathcal{U}} \sum_{t=1}^{T} 
\log p_\theta \!\left(
[\mathbf{z}^{\text{CF}}_{i_t}, \mathbf{z}^{\text{SEM}}_{i_t}]
\mid \mathbf{z}_{i_{<t}}
\right).
\]
This formulation generalizes both pure CF-based tokenization (when $g(i)=1$) and pure semantic tokenization (when $g(i)=0$),  
and lays the foundation for the proposed \emph{Dynamic Dual-Codebook Learning} framework.

\section{Methodology}

In this section, we elaborate on the proposed \textbf{FlexCode} framework. Our methodology addresses the challenges of representation entanglement and static capacity allocation in existing generative recommendation systems by introducing a novel dual codebook architecture, popularity-aware token allocation, and a robust autoregressive generation scheme. The overall framework is depicted in Figure \ref{fig:main}.

\subsection{Dual Codebook Construction}
To address the representation entanglement issue, \textbf{FlexCode} introduces two specialized codebooks for each item: a Semantic Codebook (SC) for capturing modality-specific meaning and a Collaborative Codebook (CC) for encoding high-order interaction patterns.

\subsubsection{Semantic Codebook Learning (SCL)}
The Semantic Codebook aims to distill the inherent meaning of an item from its associated textual and categorical metadata. For each item $i$, we first concatenate its descriptive attributes such as \texttt{brand}, \texttt{category}, \texttt{price}, and \texttt{title} into a unified text string. This string is then fed into a powerful pre-trained text embedding model to obtain a dense semantic embedding $\mathbf{e}_i^{\mathrm{sem}} \in \mathbb{R}^{d_{\mathrm{sem}}}$.

To convert these continuous semantic embeddings into a discrete codebook, we employ a Residual Quantization Variational Autoencoder (RQ-VAE) \cite{lee2022autoregressive}. For each $\mathbf{e}_i^{\mathrm{sem}}$, the RQ-VAE encodes it into a sequence of $L_{\mathrm{sem}}$ discrete code vectors, denoted as $\mathbf{c}_i^{\mathrm{sem}} = [\mathbf{c}_{i,1}^{\mathrm{sem}}, \ldots, \mathbf{c}_{i,L_{\mathrm{sem}}}^{\mathrm{sem}}]$, where each $\mathbf{c}_{i,j}^{\mathrm{sem}}$ is an index to an entry in a shared semantic codebook $\mathcal{V}_{\mathrm{sem}}$. The total reconstruction loss for the semantic codebook is defined as:
\begin{align}
\mathcal{L}_{\mathrm{SCL}} 
&= \mathbb{E}_{\mathbf{e}_i^{\mathrm{sem}} \sim D_{\mathrm{sem}}}
\left[ \left\| \mathbf{e}_i^{\mathrm{sem}} - 
\operatorname{Decode}\!\left(\mathbf{c}_i^{\mathrm{sem}}\right) \right\|_2^2 \right] \nonumber \\
&\quad + \sum_{j=1}^{L_{\mathrm{sem}}}
\mathcal{L}_{\mathrm{vq}}\!\left(\mathbf{z}_{i,j-1}^{\mathrm{sem}},
\mathbf{q}_{i,j}^{\mathrm{sem}}\right)
\end{align}
where $\operatorname{Decode}(\cdot)$ reconstructs the embedding from its quantized representation, and $\mathcal{L}_{\mathrm{vq}}$ is the standard VQ-VAE loss. The output of this stage for item $i$ is its semantic codebook $\mathbf{C}_{i}^{\mathrm{SEM}}$.

\subsubsection{Collaborative Codebook Learning (CCL)}
The Collaborative Codebook is designed to capture high-order co-purchase or co-view patterns. To obtain a rich collaborative embedding, we use a SASRec-like architecture \cite{kang2018self} to derive a context-aware collaborative embedding $\mathbf{e}_i^{\mathrm{col}} \in \mathbb{R}^{d_{\mathrm{col}}}$ for each item $i$. Similar to the semantic codebook, we apply an RQ-VAE to discretize these collaborative embeddings into a sequence of $L_{\mathrm{col}}$ discrete code vectors, $\mathbf{c}_i^{\mathrm{col}}$, from a collaborative codebook $\mathcal{V}_{\mathrm{col}}$. The loss function $\mathcal{L}_{\mathrm{CCL}}$ is defined analogously to $\mathcal{L}_{\mathrm{SCL}}$:
\begin{align}
\mathcal{L}_{\mathrm{CCL}}
&= \mathbb{E}_{\mathbf{e}_i^{\mathrm{col}} \sim D_{\mathrm{col}}}
\left[ \left\| \mathbf{e}_i^{\mathrm{col}} -
\operatorname{Decode}\!\left(\mathbf{c}_i^{\mathrm{col}}\right) \right\|_2^2 \right] \nonumber \\
&\quad + \sum_{j=1}^{L_{\mathrm{col}}}
\mathcal{L}_{\mathrm{vq}}\!\left(\mathbf{z}_{i,j-1}^{\mathrm{col}},
\mathbf{q}_{i,j}^{\mathrm{col}}\right)
\end{align}
The output for item $i$ is its collaborative codebook $\mathbf{C}_{i}^{\mathrm{COL}}$.

\subsection{Cross-Codebook Alignment (CCA)}
While the two codebooks capture different aspects of an item, they must not drift into unrelated representation spaces. To ensure coherence, we introduce a Cross-Codebook Alignment (CCA) objective. This objective operates on the \textit{reconstructed} embeddings, $\tilde{\mathbf{e}}_i^{\mathrm{sem}} = \operatorname{Decode}(\mathbf{C}_i^{\mathrm{SEM}})$ and $\tilde{\mathbf{e}}_i^{\mathrm{col}} = \operatorname{Decode}(\mathbf{C}_i^{\mathrm{COL}})$. Aligning the reconstructed vectors ensures that the alignment loss acts on the information that is actually preserved by the codebooks, forcing them to learn quantized representations that are not only aligned but also decodable into a coherent shared space.

Let $P_{\mathrm{sem}}(\cdot)$ and $P_{\mathrm{col}}(\cdot)$ be projection heads that map these reconstructed embeddings into a shared latent space. The alignment loss $\mathcal{L}_{\mathrm{CCA}}$ is defined using an InfoNCE objective:
\begin{equation}
\resizebox{0.9\columnwidth}{!}{$
    \mathcal{L}_{\mathrm{CCA}} = - \log 
    \frac{
        \exp\left(\text{sim}(P_{\mathrm{sem}}(\tilde{\mathbf{e}}_i^{\mathrm{sem}}), P_{\mathrm{col}}(\tilde{\mathbf{e}}_i^{\mathrm{col}})) / \tau\right)
    }{
        \sum_{j \in \mathcal{I}} 
        \exp\left(\text{sim}(P_{\mathrm{sem}}(\tilde{\mathbf{e}}_i^{\mathrm{sem}}), P_{\mathrm{col}}(\tilde{\mathbf{e}}_j^{\mathrm{col}})) / \tau\right)
    }
$}
\end{equation}
where $\text{sim}(\cdot, \cdot)$ is cosine similarity and $\tau$ is a temperature parameter. This loss pulls representations of the same item together while pushing others apart. Furthermore, this contrastive pressure serves as a powerful regularizer, encouraging more uniform usage of codebook entries and mitigating the common issue of codebook collapse in VQ-based models.

\subsection{Popularity-Aware Token Allocation (PATA)}
\label{sec:pata}
To adaptively allocate a fixed token budget $L_{\mathrm{total}}$ between the codebooks, we introduce a lightweight Mixture-of-Experts (MoE) router. The goal is to allocate more collaborative capacity to popular (head) items and more semantic capacity to sparse (tail) items.

For each item $i$, we construct a feature vector $\mathbf{x}_i = [\log(1{+}f_i), \mathrm{age}_i, \mathrm{sparsity}_i, \mathrm{uncertainty}_i]$, where $f_i$ is the normalized interaction frequency, $\mathrm{age}_i$ is the time since the item was introduced, $\mathrm{sparsity}_i$ is the inverse of its interaction density, and $\mathrm{uncertainty}_i$ is the variance of its embedding during training. While a simple sigmoid on popularity could provide a monotonic allocation, our refined MLP-based router $\phi(\cdot)$ captures more complex, non-linear interactions between these features, allowing for more nuanced allocation decisions.

The router, a shallow MLP, processes $\mathbf{x}_i$ to produce logits $[z_i^{\mathrm{col}}, z_i^{\mathrm{sem}}]$. A temperature-scaled softmax then yields the routing probabilities $\boldsymbol{\pi}_i$, from which we define the collaborative allocation ratio $\alpha_i = \pi_i^{\mathrm{col}}$. This ratio determines the soft number of tokens for each codebook:
\begin{align}
    \bar{L}_i^{\mathrm{col}} &= \alpha_i L_{\mathrm{total}}, &
    \bar{L}_i^{\mathrm{sem}} &= (1 - \alpha_i) L_{\mathrm{total}}.
\end{align}
To maintain differentiability, we use sigmoid-based masks to softly select tokens during training:
\begin{align}
    m_{k}^{\mathrm{col}}(\alpha_i) &= \sigma\!\left(\frac{\bar{L}_i^{\mathrm{col}} - (k - \tfrac{1}{2})}{\tau_m}\right), \\
    m_{k}^{\mathrm{sem}}(\alpha_i) &= \sigma\!\left(\frac{\bar{L}_i^{\mathrm{sem}} - (k - \tfrac{1}{2})}{\tau_m}\right),
\end{align}
where the hyperparameter $\tau_m$ controls the smoothness of the soft-to-hard transition. We set $\tau_m=0.1$ and found the model to be robust to small variations of this value. The final combined representation $\mathbf{C}_i$ concatenates the masked codebooks. At inference, the allocation is discretized via rounding. To ensure router stability, we use stratified load-balancing ($\mathcal{L}_{\mathrm{lb}}$) within popularity bands and a local smoothness regularizer ($\mathcal{L}_{\mathrm{smooth}}$) on the allocation ratios $\alpha_i$.

\subsection{Autoregressive Generation (ARG)}
Given a user's historical sequence of items $S_u = (i_1, \ldots, i_{T-1})$, we retrieve their corresponding combined codebooks $(\mathbf{C}_{i_1}, \ldots, \mathbf{C}_{i_{T-1}})$. An autoregressive Transformer model is trained to maximize the likelihood of the next item's codebook $\mathbf{C}_{i_T}$. The loss is the standard cross-entropy over the sequence of predicted tokens:
\begin{equation}
\begin{split}
    \mathcal{L}_{\mathrm{ARG}} = 
    - \sum_{t=1}^{N_{\mathrm{seq}}-1} \sum_{k=1}^{L_{\mathrm{total}}} 
    \log P(&\mathbf{c}_{i_{t+1},k} \,|\, 
    \mathbf{C}_{S_u^{<t+1}}, \\
    &\qquad \mathbf{c}_{i_{t+1}, <k})
\end{split}
\end{equation}
where $\mathbf{C}_{S_u^{<t+1}}$ represents the context from previous items and $\mathbf{c}_{i_{t+1}, <k}$ represents previously generated tokens for the current target item.

\subsection{Overall Objective and Training}
The entire \textbf{FlexCode} framework is trained jointly in an end-to-end manner. The final objective function is a weighted sum of all components: the codebook reconstruction losses, the cross-codebook alignment loss, the autoregressive generation loss, and the routing regularization losses.
\begin{align}
\mathcal{L}_{\mathrm{total}} ={}& \mathcal{L}_{\mathrm{SCL}} + \mathcal{L}_{\mathrm{CCL}} 
+ \lambda_{\mathrm{CCA}}\mathcal{L}_{\mathrm{CCA}} \nonumber \\
& + \lambda_{\mathrm{ARG}}\mathcal{L}_{\mathrm{ARG}}
+ \lambda_{\mathrm{lb}}\mathcal{L}_{\mathrm{lb}}
+ \lambda_{\mathrm{smooth}}\mathcal{L}_{\mathrm{smooth}}
\end{align}
where $\lambda_{(\cdot)}$ are hyperparameters that balance the contribution of each term. This end-to-end training allows the codebooks, the router, and the autoregressive model to co-adapt, leading to a more effective and cohesive final system.

\section{Experiments}

In this section, we empirically demonstrate the effectiveness and efficiency of the proposed method \textbf{FlexCode}.

\subsection{Experimental Setup}
\subsubsection{Datasets}

\begin{table*}[t]
\centering
\caption{
Overall comparison of \textbf{FlexCode} and baseline models on three datasets.
Best results are in \textbf{bold}, and second-best are \underline{underlined}. All improvements are statistically significant with $p < 0.01$.
}
\label{tab:three-datasets}
\small
\setlength{\tabcolsep}{5pt}
\scalebox{0.9}{
\begin{tabular}{lcccccccccccc}
\toprule
\multirow{2}{*}{\textbf{Model}} &
\multicolumn{4}{c}{\textbf{Beauty}} &
\multicolumn{4}{c}{\textbf{Sports and Outdoors}} &
\multicolumn{4}{c}{\textbf{KuaiRand}} \\
\cmidrule(lr){2-5} \cmidrule(lr){6-9} \cmidrule(lr){10-13}
& R@5 & N@5 & R@10 & N@10
& R@5 & N@5 & R@10 & N@10
& R@5 & N@5 & R@10 & N@10 \\
\midrule
\multicolumn{13}{c}{\emph{Item ID-based methods}} \\
\midrule
Caser     & 0.0205 & 0.0131 & 0.0347 & 0.0176 & 0.0116 & 0.0072 & 0.0194 & 0.0097 & 0.0074 & 0.0068 & 0.0118 & 0.0095 \\
GRU4Rec   & 0.0164 & 0.0099 & 0.0283 & 0.0137 & 0.0129 & 0.0086 & 0.0204 & 0.0110 & 0.0298 & 0.0217 & 0.0383 & 0.0245 \\
HGN       & 0.0325 & 0.0206 & 0.0512 & 0.0266 & 0.0189 & 0.0120 & 0.0313 & 0.0159 & 0.0297 & 0.0169 & 0.0354 & 0.0219 \\
BERT4Rec  & 0.0203 & 0.0124 & 0.0347 & 0.0170 & 0.0115 & 0.0075 & 0.0191 & 0.0099 & 0.0185 & 0.0196 & 0.0217 & 0.0236 \\
SASRec    & 0.0387 & 0.0249 & 0.0605 & 0.0318 & 0.0233 & 0.0154 & 0.0350 & 0.0192 & 0.0332 & 0.0338 & 0.0405 & 0.0372 \\
S$^3$-Rec & 0.0387 & 0.0244 & 0.0647 & 0.0327 & 0.0251 & 0.0161 & 0.0385 & 0.0204 & -- & -- & -- & -- \\
Recformer & 0.0379 & 0.0257 & 0.0589 & 0.0321 & 0.0249 & 0.0154 & 0.0370 & 0.0201 & -- & -- & -- & -- \\
\midrule
\multicolumn{13}{c}{\emph{Semantic ID-based methods}} \\
\midrule
VQ-Rec    & 0.0457 & 0.0317 & 0.0664 & 0.0383 & 0.0208 & 0.0144 & 0.0300 & 0.0173 & 0.0513 & 0.0354 & 0.0589 & 0.0412 \\
TIGER     & 0.0454 & 0.0321 & 0.0648 & 0.0384 & 0.0264 & 0.0181 & 0.0400 & 0.0225 & 0.0557 & 0.0383 & 0.0624 & 0.0445 \\
LC-Rec    & 0.0478 & 0.0329 & 0.0679 & 0.0389 & 0.0268 & 0.0177 & 0.0412 & 0.0221 & 0.0622 & 0.0403 & 0.0684 & 0.0497 \\
COBRA     & 0.0537 & 0.0395 & 0.0725 & 0.0456 & 0.0306 & 0.0215 & 0.0434 & 0.0257 & -- & -- & -- & -- \\
URL       & \underline{0.0553} & \underline{0.0410} & \underline{0.0736} & \underline{0.0471} & \underline{0.0305} & \underline{0.0218} & \underline{0.0449} & \underline{0.0273} & \underline{0.0654} & \underline{0.0481} & \underline{0.0778} & \underline{0.0585}  \\
\midrule
FlexCode-SID only & 0.0510 & 0.0375 & 0.0689 & 0.0433 & 0.0291 & 0.0204 & 0.0412 & 0.0244 & 0.0523 & 0.0461 & 0.0759 & 0.0517\\
FlexCode-CF only  & 0.0360 & 0.0232 & 0.0563 & 0.0296 & 0.0217 & 0.0143 & 0.0326 & 0.0179 & 0.0309 & 0.0314 & 0.0377 & 0.0346 \\
FlexCode-Fix      & 0.0531 & 0.0394 & 0.0707 & 0.0452 & 0.0293 & 0.0209 & 0.0431 & 0.0262 & 0.0628 & 0.0462 & 0.0747 & 0.0562 \\
\textbf{FlexCode (ours)} & 
\textbf{0.0578} & \textbf{0.0415} & \textbf{0.0769} & \textbf{0.0483} &
\textbf{0.0329} & \textbf{0.0232} & \textbf{0.0471} & \textbf{0.0275} &
\textbf{0.0709} & \textbf{0.0524} & \textbf{0.0825} & \textbf{0.0632}\\
\bottomrule
\end{tabular}}
\end{table*}

To ensure a comprehensive evaluation of our framework across different domains and data characteristics, we conduct experiments on three widely used public benchmarks and a large-scale proprietary industrial dataset. Following prior work~\citep{he2017neural, sun2019bert4rec, zhou2020s3}, we treat each user’s chronological interaction history as an ordered sequence. We adopt the standard \emph{leave-last-out} evaluation protocol~\citep{kang2018self}, using the last item for testing and the second-to-last for validation. Following common practice, we apply the 5-core filtering strategy.

\paragraph{Public Benchmarks}
We evaluate \textbf{FlexCode} on three public datasets: \textbf{Amazon-Beauty}~\citep{mcauley2015image}, a dense, medium-scale e-commerce dataset; \textbf{Amazon-Sports and Outdoors}~\citep{mcauley2015image}, a larger and sparser e-commerce dataset; and \textbf{KuaiRand-1K}~\citep{gao2022kuairand}, a large-scale short-video recommendation dataset with rich side information. Detailed statistics are in Table~\ref{tab:dataset-stats}.

\begin{table}[h]
\centering
\caption{Statistics of public benchmark datasets after preprocessing.}
\scalebox{0.8}{
\begin{tabular}{lcccc}
\toprule
Dataset & \#Users & \#Items & \#Interactions & Avg. Seq.\\
\midrule
Beauty & 22{,}363 & 12{,}101 & 198{,}360 & 8.87 \\
Sports & 35{,}598 & 18{,}357 & 296{,}175 & 8.32 \\
KuaiRand & 1{,}000 & 3.6M & 11M & 11.71 \\
Proprietary & 1.5M+ & 1M+ & 45M+ & 28.0\\
\bottomrule
\end{tabular}}
\label{tab:dataset-stats}
\end{table}

\paragraph{Proprietary Industrial Dataset}
To further examine the effectiveness of FlexCode under production-scale conditions, we additionally train and evaluate our model on an in-house industrial dataset collected from a commercial platform. The dataset contains tens of millions of users and their interaction sequences over tens of thousands of items, recorded over a period exceeding one year. We use the last day of logged interactions for evaluation to reflect an online inference scenario. Due to confidentiality agreements, we report only approximate, rounded statistics in Table~\ref{tab:dataset-stats}.
\subsubsection{Baselines}
\textbf{Item ID-based methods}
We include widely used sequence models that learn item representations directly from user--item interaction histories:
Caser~\citep{tang2018personalized},
GRU4Rec~\citep{hidasi2015session},
HGN~\citep{ma2019hierarchical},
BERT4Rec~\citep{sun2019bert4rec},
SASRec~\citep{kang2018self},
Recformer~\citep{li2023text},
and S3-Rec~\citep{zhou2020s3}.
\noindent\textbf{Semantic ID-based methods}
We further compare with recent approaches that tokenize or quantize item representations into discrete semantic identifiers:
VQRec~\citep{hou2023learning},
TIGER~\citep{rajput2023recommender},
LC-Rec~\citep{zheng2024adapting}, COBRA \cite{yang2025sparse}
and Unified Representation Learning~\citep{lin2025unified}.

\subsubsection{Evaluation Metrics}
We assess model performance using \textbf{Recall@K} and \textbf{NDCG@K}, with $K \in \{5, 10\}$, following prior work~\citep{rajput2023recommender}.

\subsection{Main Results on Public Benchmarks}
Table~\ref{tab:three-datasets} presents the overall performance comparison on the three public benchmarks, where \textbf{FlexCode} consistently outperforms all baseline models across all metrics. The advantage is most pronounced on the larger and sparser datasets. On Amazon-Sports, \textbf{FlexCode} achieves a Recall@10 of 0.0471, a 5.3\% relative improvement over the strongest semantic baseline, URL. This performance gap widens on the large-scale KuaiRand dataset, where \textbf{FlexCode} attains an NDCG@10 of 0.0632, representing an 8.0\% improvement over URL and demonstrating our model's robustness to diverse data characteristics. This consistent superiority stems from its architecture, which directly addresses the limitations of prior methods. Unlike Item ID-based models such as SASRec, which lack a mechanism for semantic generalization on sparse data, \textbf{FlexCode} utilizes a dedicated semantic codebook. At the same time, it avoids the representation entanglement common in single-codebook generative models like URL by explicitly disentangling collaborative and semantic signals into separate codebooks.

The value of this design is further substantiated by the performance of our model's ablated variants. The \textbf{FlexCode-Fix} version, which uses a static 50/50 token split, already outperforms most baselines, confirming the inherent benefit of the disentangled dual-codebook structure. However, the full \textbf{FlexCode} model consistently outperforms \textbf{FlexCode-Fix}, isolating the critical contribution of the popularity-aware token allocation (PATA) mechanism. For example, on KuaiRand, the dynamic allocation of PATA lifts NDCG@10 from 0.0562 to 0.0632, a relative gain of 12.5\% over the fixed-split model. This confirms that both architectural components—disentanglement and adaptive allocation—are essential to achieving state-of-the-art performance.

\subsection{Further Analysis on Industrial Dataset}
After establishing \textbf{FlexCode}'s overall superiority, we conduct analyses on our proprietary industrial dataset to examine its robustness and adaptability under production-scale conditions.

\begin{figure*}[t]
\centering
\begin{subfigure}[t]{0.48\linewidth}
\vspace{0pt}
\centering
\includegraphics[width=\linewidth]{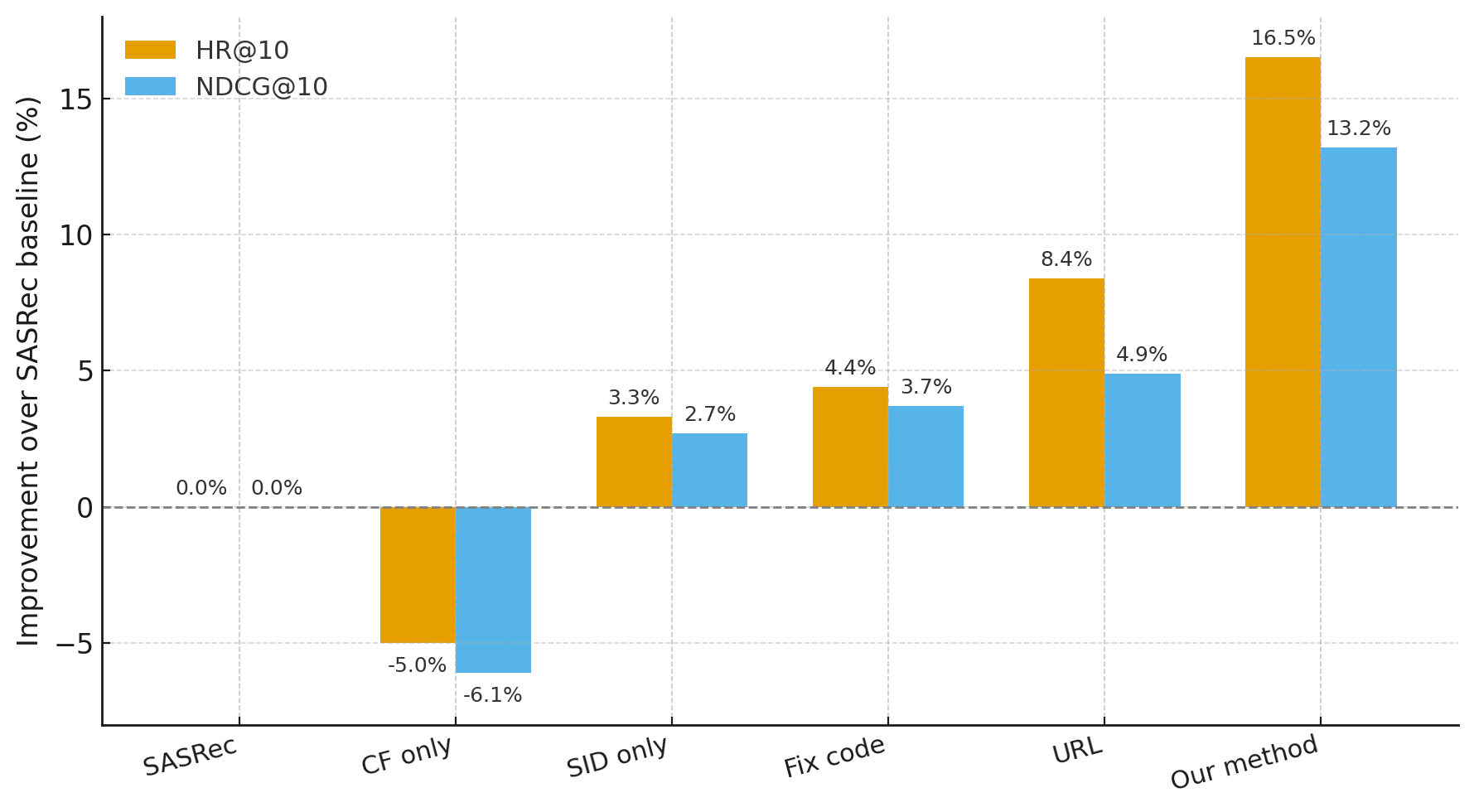}
\caption{Overall performance improvement of various models over the SASRec baseline on the industrial dataset.}
\label{fig:industry-main}
\end{subfigure}
\hfill
\begin{subfigure}[t]{0.48\linewidth}
\vspace{0pt}
\centering
\includegraphics[width=\linewidth]{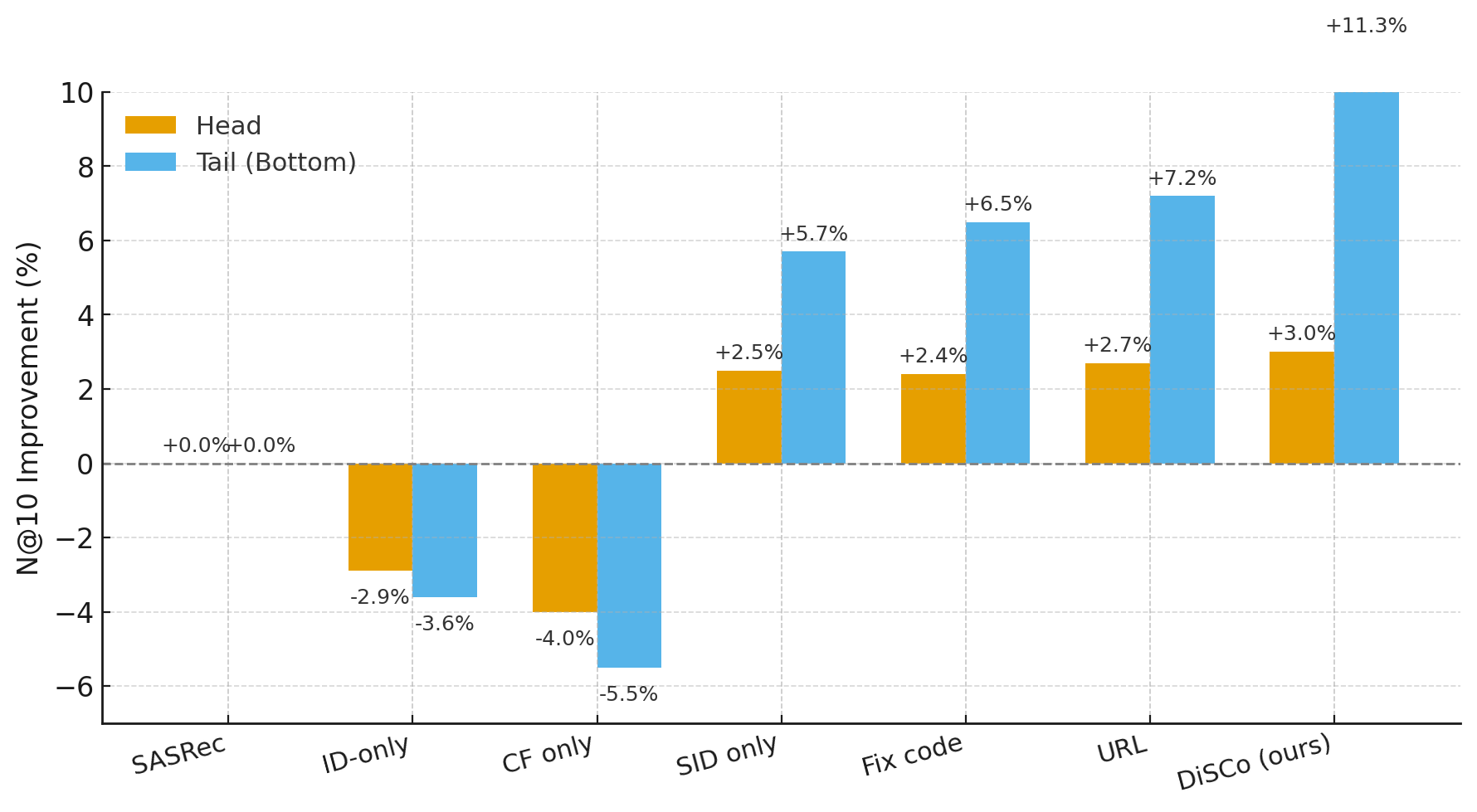}
\caption{NDCG@10 improvement on Head vs. Tail items.}
\label{fig:headtail}
\end{subfigure}
\caption{Performance evaluation on the large-scale industrial dataset.}
\end{figure*}

\subsubsection{Overall Industrial Performance}
Figure~\ref{fig:industry-main} shows the performance improvement of generative models relative to SASRec baseline. While a pure collaborative generative model (CF only) degrades performance by over 5\% and advanced unified models like URL provide a moderate 4.9\% lift in NDCG@10, \textbf{FlexCode} delivers a 13.2\% improvement in NDCG@10 and a 16.5\% improvement in HR@10. This large margin of victory on a massive, noisy, and production-level dataset confirms that the architectural principles of \textbf{FlexCode} are not only theoretically sound but also scalable and highly effective in a real-world industrial setting.

\subsubsection{Cold-Start and Long-Tail Item Performance}
A central claim of this work is that adaptive allocation is critical for balancing memorization for popular items and generalization for rare items. We test this hypothesis by measuring NDCG@10 on the \textbf{Head} and \textbf{Bottom(Tail)} items from our industrial dataset, which has a highly skewed long-tail distribution. The results in Figure~\ref{fig:headtail} provide direct evidence for our claim. Baselines exhibit a clear performance trade-off: the CF-only model improves head performance but degrades tail performance by 5.5\%, whereas the SID-only model improves the tail by 5.7\% at the cost of weaker head performance. \textbf{FlexCode} resolves this conflict. For head items, its MoE router correctly allocates more tokens to the collaborative codebook, enabling fine-grained memorization and yielding a 3.0\% NDCG@10 improvement. For tail items, the router shifts capacity to the semantic codebook to enable generalization from content features, resulting in an 11.3\% NDCG@10 improvement. This result is the largest gain on tail items by a wide margin and is of high practical importance, as improving long-tail discovery is a primary goal of industrial recommender systems. By dynamically allocating its representational budget, \textbf{FlexCode} demonstrates superior performance across the entire item popularity spectrum.

\section{Ablation Study}
We conduct ablation experiments on the KuaiRand dataset to analyze the contribution of each component in \textbf{FlexCode}. Table~\ref{tab:ablation_revised} summarizes the results. Removing the dual-codebook structure by using only collaborative IDs (\emph{CID Only}) or semantic IDs (\emph{SID Only}) leads to a significant performance drop, showing that separating and combining collaborative and semantic representations is critical. Disabling the MoE gating network and using a fixed 50/50 split (\emph{w/o MoE Gating}) reduces adaptability to item popularity and harms performance. Finally, excluding the alignment loss (\emph{w/o Alignment Loss}) causes a drop in performance, confirming its importance for maintaining cross-codebook coherence.

\begin{table}[h!]
\centering
\caption{Ablation study on the KuaiRand dataset. Each row removes or modifies a key component of \textbf{FlexCode}.}
\label{tab:ablation_revised}
\scalebox{0.75}{
\begin{tabular}{lcc}
\toprule
\textbf{Model Variant} & \textbf{Recall@10} & \textbf{NDCG@10} \\
\midrule
\textbf{FlexCode (Full)} & \textbf{0.0825} & \textbf{0.0632} \\
FlexCode (CID Only) & 0.0405 & 0.0372 \\
FlexCode (SID Only) & 0.0511 & 0.0401 \\
w/o MoE Gating (Fixed Split) & 0.0791 & 0.0598 \\
w/o Alignment Loss & 0.0809 & 0.0615 \\
\bottomrule
\end{tabular}}
\end{table}

\paragraph{Token Budget Sensitivity.}
We further analyze \textbf{FlexCode}'s robustness under varying token budgets $L \in \{3,4,5,6\}$. As shown in Table~\ref{tab:budget_sensitivity}, \textbf{FlexCode} consistently achieves strong performance even with reduced token capacity. This demonstrates that its adaptive capacity reallocation allows for more efficient use of limited representational resources, a key advantage for real-world deployment.

\begin{table}[h!]
\centering
\caption{Token budget sensitivity on the KuaiRand dataset (NDCG@10).}
\label{tab:budget_sensitivity}
\scalebox{0.75}{
\begin{tabular}{lcccc}
\toprule
\textbf{Model} & \textbf{L = 3} & \textbf{L = 4} & \textbf{L = 5} & \textbf{L = 6} \\
\midrule
FlexCode (SID Only) & 0.0401 & 0.0415 & 0.0418 & 0.0420 \\
FlexCode (CID Only) & 0.0372 & 0.0389 & 0.0395 & 0.0397 \\
FlexCode-Fix (50/50 Split) & 0.0598 & 0.0615 & 0.0619 & 0.0621 \\
\textbf{FlexCode (ours)} & \textbf{0.0632} & \textbf{0.0685} & \textbf{0.0691} & \textbf{0.0693} \\
\bottomrule
\end{tabular}}
\end{table}

\paragraph{Hyper-Parameter Analysis.}
We examine the effect of key hyper-parameters in Table~\ref{tab:hyper}. \textbf{FlexCode} is generally stable across a broad range of settings. Larger codebooks ($K$) and dimensions ($d$) offer slight improvements until saturation. Performance is robust to moderate changes in the regularization weights ($\lambda_{\text{align}}$, $\lambda_{\text{smooth}}$), suggesting that the model is not overly sensitive to precise tuning.

\begin{table}[t]
\centering
\caption{Hyper-parameter sensitivity analysis on the KuaiRand dataset.}
\label{tab:hyper}
\scalebox{0.65}{
\begin{tabular}{lccccc}
\toprule
\textbf{Parameter Setting} & \textbf{$K$} & \textbf{$d$} & \textbf{$\lambda_{\text{align}}$} & \textbf{$\lambda_{\text{smooth}}$} & \textbf{NDCG@10} \\
\midrule
\textbf{Default (Base)} & \textbf{512} & \textbf{64} & \textbf{0.1} & \textbf{0.01} & \textbf{0.0632} \\
\midrule
\textit{K Variation} & 256 & 64 & 0.1 & 0.01 & 0.0603 \\
& 1024 & 64 & 0.1 & 0.01 & 0.0635 \\
\midrule
\textit{d Variation} & 512 & 32 & 0.1 & 0.01 & 0.0611 \\
& 512 & 128 & 0.1 & 0.01 & 0.0639 \\
\midrule
\textit{$\lambda_{\text{align}}$ Variation} & 512 & 64 & 0.01 & 0.01 & 0.0618 \\
& 512 & 64 & 0.5 & 0.01 & 0.0625 \\
& 512 & 64 & 1.0 & 0.01 & 0.0609 \\
\midrule
\textit{$\lambda_{\text{smooth}}$ Variation} & 512 & 64 & 0.1 & 0.001 & 0.0621 \\
& 512 & 64 & 0.1 & 0.05 & 0.0628 \\
& 512 & 64 & 0.1 & 0.1 & 0.0615 \\
\bottomrule
\end{tabular}}
\end{table}
\section{Conclusion}
\label{sec:conclusion}

In this work, we approached generative recommendation from the perspective of \emph{adaptive capacity allocation}, introducing \textsc{FlexCode}, a dual-codebook framework with popularity-aware routing that improves both overall accuracy and long-tail performance.
More broadly, our results suggest several promising directions for future work.
One avenue is to extend adaptive capacity allocation to richer multi-modal item descriptors and to user-side modeling, enabling joint reasoning over user and item codebooks.
Another is to integrate \textsc{FlexCode} with large language models and online learning pipelines, studying how popularity-aware tokenization interacts with exploration, temporal drift, and calibration in real-world deployments.
It is also important to examine fairness and exposure of long-tail content under different routing strategies, and to develop theoretical tools for understanding when and why dual-codebook architectures provide benefits over unified tokenizations.
We hope that viewing generative recommendation through the lens of dual codebooks and adaptive capacity allocation will inspire further work on balancing memorization and generalization in token-based recommender systems.


\bibliography{custom}
\clearpage
\appendix

\end{document}